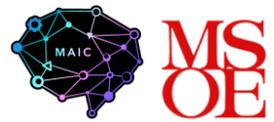

# Machine Learning-Assisted Vocal Cord Ultrasound Examination: Project VIPR

Will Sebelik-Lassiter, Evan Schubert, Muhammad Alliyu, Quentin Robbins, Excel Olatunji, Mustafa Barry

Department Electrical, Computer and Biomedical Engineering

Department of Computer Science and Software Engineering

Milwaukee School of Engineering

1025 N Broadway St, Milwaukee, WI 53202

[lassiterw@msoe.edu](lassiterw@msoe.edu)

Merry Sebelik, MD

Department of Otolaryngology, Head & Neck Surgery

Emory University

550 Peachtree Ave NE, Suite 1135, Atlanta, GA 30308

[msebeli@emory.edu](msebeli@emory.edu)

## Abstract

Intro: Vocal cord ultrasound (VCUS) has emerged as a less invasive and better tolerated examination technique, but its accuracy is operator dependent. This research aims to apply a machine learning-assisted algorithm to automatically identify the vocal cords and distinguish normal vocal cord images from vocal cord paralysis (VCP).

Methods: VCUS videos were acquired from 30 volunteers, which were split into still frames and cropped to a uniform size. Healthy and simulated VCP images were used as training data for vocal cord segmentation and VCP classification models.

Results: The vocal cord segmentation model achieved a validation accuracy of 96%, while the best classification model (VIPRnet) achieved a validation accuracy of 99%.

Conclusion: Machine learning-assisted analysis of VCUS shows great promise in improving diagnostic accuracy over operator-dependent human interpretation.

# 1. Introduction

Vocal cord paralysis (VCP) refers to a condition in which one or both vocal cords (also known as vocal folds) are unable to move, leading to significant health impacts. VCP can occur as a result of surgery (especially thyroid, parathyroid or neck surgeries), trauma, tumors, or neurological disorders. The inability of the vocal cords to move can lead to a weak or breathy voice, hoarseness, swallowing dysfunction, and in severe cases, difficulty breathing.

Examination methods to visualize vocal cords have been developed over many decades. Benjamin Guy Babington in 1829 is credited with being the first observer of the vocal cords, using a mirror and reflected light [1]. Today, flexible laryngoscopy (FL) is considered the current gold standard for examination of the vocal cords in an awake patient [2]. This technique is unfortunately an invasive one, as it involves navigating a flexible camera probe through one's mouth into the upper parts of the throat. Additionally, some types of probes are discarded after a single use, while the reusable ones must undergo a lengthy sterilization process after each use. The less invasive method of transcutaneous ultrasound of the vocal cords (VCUS) has gained acceptance due to lower cost, improved patient tolerance, reduced risk of respiratory infectious exposure, and accuracy in both pediatric and adult patients [3,4].

VCUS can be particularly suited to the pediatric population [5]. Hamilton, et al. conducted a systematic review and meta-analysis demonstrating that VCUS has high sensitivity (91%) and specificity (97%) for detecting vocal cord immobility in children, making it a reliable alternative to laryngoscopy in pediatric populations [6]. Similarly, Su et al. found that VCUS in adults has a pooled sensitivity of 95% and specificity of 99% for diagnosing true vocal fold immobility, supporting its utility in adult patients as well [7].

Despite the apparent advantages of VCUS in terms of cost and patient tolerance, it is recognized that the examination technique can be highly operator-dependent, with a range of vocal cord identification from 41% to 86% per a systematic review by Carneiro-Pla, et al in 2014 [8]. Further, some authors have recommended against using VCUS to identify VCP especially in overweight and postoperative patients [9].

To overcome operator-dependence, machine learning has been applied to vocal cord motion analysis, as exemplified in the 2010 work by Voight, et al. who used phonovibrogram features and various machine learning (ML) algorithms to achieve classification accuracies of up to 93% [10]. More recently, deep learning methods have been used to analyze video laryngoscopy images obtained from FL [11, 13]. Application of machine learning to VCUS interpretation has not yet been described.

This research aims to apply a machine learning-assisted algorithm to automatically identify and segment human vocal cords from ultrasound images, and investigates the application of multiple deep learning algorithms to differentiate bilaterally mobile vocal cords from VCP. This was done with the intent of complementing clinician examination and interpretation.

# 2. Methods

As of March 2025, we were unable to identify any publicly available datasets of significance on the internet which contained ultrasound images or videos of the vocal cords. Thus, approval from the Milwaukee School of Engineering Institutional Review Board (MSOE IRB# I-2024-012) was obtained for a study which would allow us to record videos of VCUS scans from consenting participants and train a machine learning-based vocal cord segmentation model on the videos. Over the course of two weeks, 30 MSOE undergraduate students were recruited to provide ultrasound videos of their vocal cords. Exclusion criteria ensured that all study participants were "MSOE community members" (students, faculty, or staff) and excluded those who anticipated a possible inflammatory reaction to ultrasound gel or its ingredients. Paper flyers posted in common areas around MSOE campus were used to recruit participants and free pizza was offered as a participation incentive. Each participant verbally consented after an investigator read them an informed consent script that detailed the specifics of the study. Precautions were taken to ensure that all collected data remained de-identified and confidential. No identifying information was collected from any participant except self-identified gender and age.

| Number of Male Participants | 7 |
|---|---|
| Number of Female Participants | 23 |
| Median Age | 20 |

Table 1: Participant demographic breakdown

Each participant was comfortably seated for the ultrasound examiner to record a 30 to 60 second video showing their vocal cords. Care was taken to include all pertinent anatomy necessary to fully visualize vocal motion. Anatomic structures typically seen on VCUS are illustrated in Figure 1. Participants were observed during passive respiration and during phonation, throat clearing and laughter. Ultrasound videos were recorded using a GE Logiq S7 Pro ultrasound machine and linear probe at 8.5 MHz, depth 5cm, using a standardized examination technique. A single author (WSL) was trained and certified in VCUS, therefore he carried out all data acquisition. To preserve participant anonymity, all personally identifiable information, including overlaid text within ultrasound frames, was removed using a custom anonymization script.

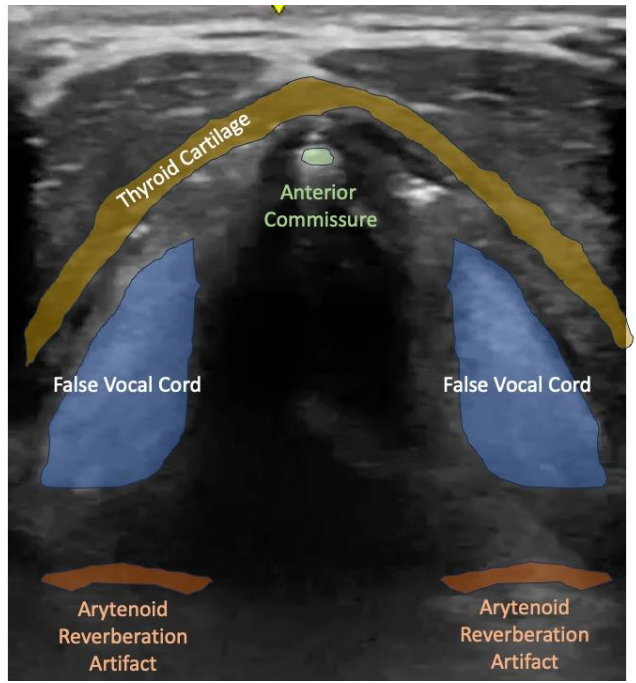

Figure 1: Vocal cord ultrasound image with annotated pertinent anatomy

## 2.1 Data Processing and Labeling

All collected ultrasound videos were securely uploaded over a secure Wi-Fi network to MSOE's on-campus supercomputer, Rosie. Three participant videos (10% of overall dataset) were moved to a separate directory to serve as validation data, and one video was excluded entirely due to its low imaging quality. The remaining 26 videos were processed to extract every 20$^{th}$ frame, resulting in a total of 2,168 still frames parsed from video. Next, each extracted frame was resized to 256×256 pixels via Lanczos interpolation and saved in full-color PNG format to maintain a consistent dimension and file type. Lastly, an anonymization script was run on each frame which removed any remaining text or scale bars from the images, further simplifying and standardizing the dataset.

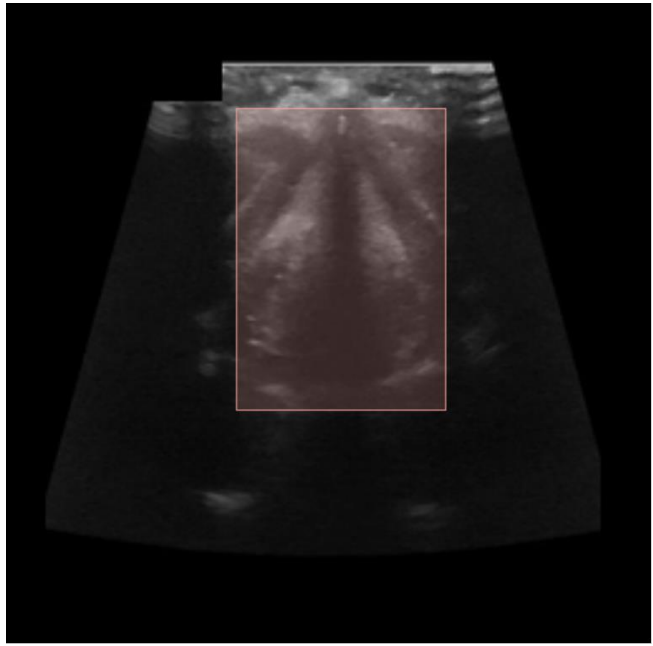

Figure 2: Example of ROI annotation/labeling on an anonymized VCUS video frame

After cleaning and resizing, these images were set aside to have the vocal cords segmented manually by three team members (WSL, QR, MB), using the Label Studio open-source software run locally on Rosie. Each annotator identified a rectangular region of interest (ROI) by placing the upper boundary of the region at the anterior commissure (crest where the two cords meet), the lower boundary at the highest visible reverberation artifacts from the arytenoid cartilage, and the left and right boundaries directly adjacent to the visible edges of the vocal cords (Figure 2). Frames that lacked clearly visible vocal cords were discarded, leaving us with a model training dataset totaling 1,088 images and 1,088 associated ROI labels (one per image).

## 2.2 Segmentation Model Training

These 1,088 manually labeled images were used to train a segmentation model which can identify and segment the vocal cords in a VCUS image, if they are sufficiently visible. We used the YOLOv8m object detection model provided by Ultralytics as the backbone for this task, benefiting from its efficient convolutional architecture and decoupled detection head. All experiments were conducted using ultralytics==8.3.88.

Images were resized to 640×640 during model training. The training was performed with augment=True enabled in YOLOv8, which includes default augmentations such as mosaic, mixup, scaling, and HSV color-space augmentation. The model was trained for 4 epochs using the SGD optimizer with the default learning rate, momentum (0.937), and weight decay (5e-4).

Training and validation sets were stratified from the augmented dataset. The model was trained on a single-class detection task ("vocal_cords") using YOLO's standard CIoU loss for bounding box regression, Binary Cross-Entropy for classification, and Distribution Focal Loss (DFL) for refined localization.

## 2.3 Segmentation Model Evaluation Protocol

We used COCO-style object detection metrics to evaluate the model on the 10% validation set drawn exclusively from the three participant videos that were previously unseen by any model:

- Precision, Recall, and F1-score
- mAP@0.5 and mAP@0.5:0.95
- Confidence-curve and PR-curve analysis
- Normalized confusion matrix (vocal cords vs. background)

The model was evaluated on a validation set using a confidence threshold of 0.701, derived from the F1-Confidence curve. While clinical applications often prioritize precision to minimize false positives, F1-score was selected here to balance the high recall needed for screening and early detection tasks with acceptable precision.

## 2.4 Preparation for VCP Classification

After building a vocal cord detection and segmentation model, we then sought to investigate the viability of using a convolutional neural network (CNN) to identify potential vocal cord paralysis from ultrasound images. We investigated two different CNN architectures to perform a binary classification task on input VCUS images: differentiating between bilaterally mobile vocal cords and cords exhibiting VCP. To address the major limitation of not having access to a source of clinical VCUS images of patients with VCP, we were able to create digitally simulated images of VCP. Vocal cord length is noted to vary by functional activity such as singing at high or low pitch [12], or clinical conditions such as paralysis. VCP causes distinctive vocal cord length difference between the paralyzed side and the mobile side [2], thus inspiring and supporting our method of digital alteration to simulate VCP.

### 2.4.1 Synthetic Data Creation and Data Augmentation

To address class imbalance when approaching classification, we employed an "asymmetric squishing" technique to create examples of VCP from the 1,088 healthy training images we had available to us. First, we referenced our manually created labels to crop each image down to just the ROI, so that each image exclusively contained vocal cord anatomy, and then we resized each image back up to 256x256 using Lanczos interpolation. Each ROI image was then split in half vertically down the center of the image, and each half was saved separately for the subsequent synthetic data creation steps.

To simulate the appearance of one paralyzed vocal cord being shorter than the other (mobile) cord, one of the image halves was scaled vertically by a factor of 0.75 in such a way that the top borders of the two half-images stayed lined up, but the bottom border of the "compressed" half now rested well above the bottom border of the unmodified half. Pixel information was then sampled from the original VCUS image, i.e. from outside of the labeled ROI, to fill in the "empty space" below this compressed half and make the image square again. This created the appearance that the vocal cord contained in one half of the image was noticeably shorter than the vocal cord in the other half.

Unfortunately, recomposing the compressed half with the uncompressed half often resulted in a "seam" where the pixels no longer lined up neatly. To address this, a vertical strip six pixels wide on either side of this seam line was deleted and replaced using a bilinear interpolation based on the brightness values of the pixels adjacent to the seam. This looked more natural, but unfortunately still could be considered a significant visual artifact in many of our training images. We were concerned that this artifact would induce unintended bias in our classification models if it were present only in our synthetic VCP example images, so we came up with a strategy that we hoped would mitigate this issue and simultaneously ensure that our dataset contained an equal distribution of both healthy and paralyzed training images:

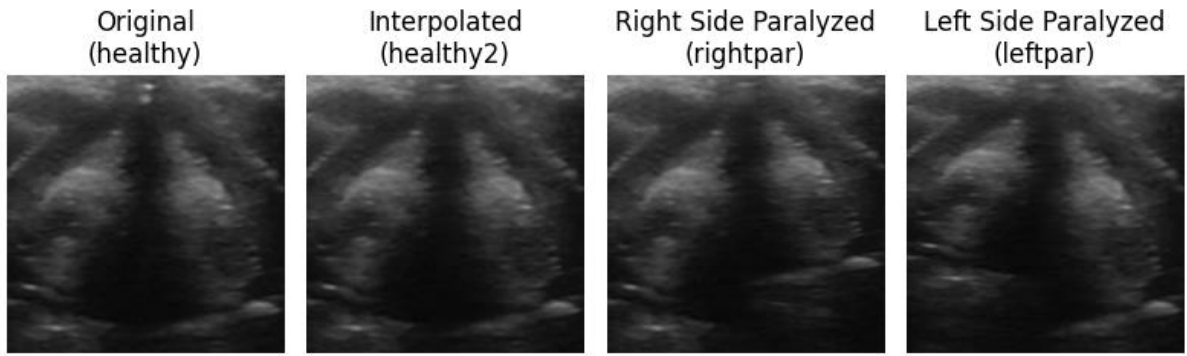

Figure 3: Examples of the four classes of images created

Four image groups were created for our training data. The first group, "rightpar", contained 1,088 images that had been modified by the procedure described above to have their right sides compressed in a simulacrum of right vocal fold immobility. Similarly, "leftpar" contained 1,088 more images which had their left halves compressed equivalently. In an attempt to reduce artifact bias, the third image group called "healthy2" contained 1,088 more images that were split vertically and then rejoined using the same bilinear interpolation operation as had been used on the paralysis images, but neither half was compressed or modified otherwise. This way, the image still looked mostly "normal", but exhibited the same bilinear interpolation artifact present in the paralysis images. Finally, the fourth group contained a final 1,088 images that were not modified beyond cropping them down to the labelled ROI. After creating these four groups, we had 2,176 healthy examples and 2,176 VCP examples, or 4,352 training images in total.

Finally, data augmentation was applied to both healthy and VCP image groups to further expand the dataset and enhance model generalizability. Augmentations included random rotations of up to 10 degrees, horizontal flips, and minor affine transformations. In addition, multiple augmentations were combined sequentially, such as a rotation followed by a flip, or an affine transformation followed by both a rotation and a flip, to introduce additional variability. This approach substantially increased the number of training-ready images, raising the total to 34,816 unique images.

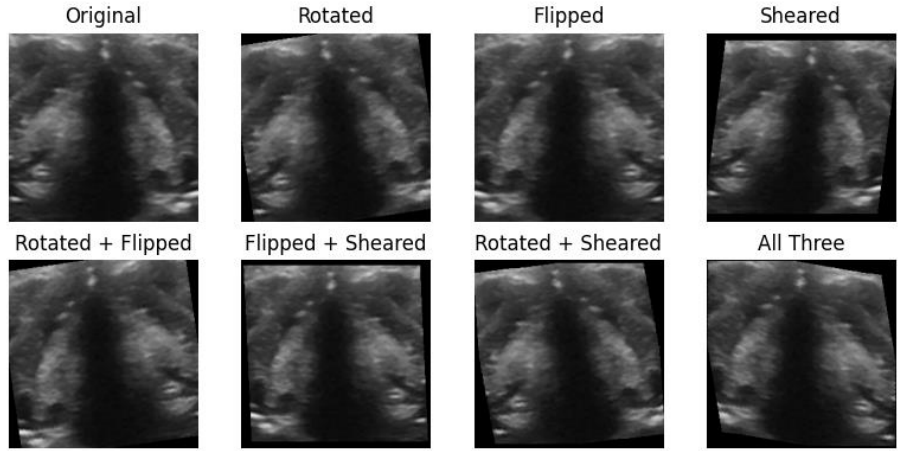

Figure 4: Examples of how original images were augmented to increase dataset size

## 2.5 Classification Models

We evaluated the performance of two different CNN architectures for classifying VCUS images as healthy or paralyzed. The first model we evaluated, YOLOv8n-cls, was pre-trained on large-scale image datasets and then fine-tuned on our augmented VCUS dataset. Then, we developed and evaluated a simple custom model architecture which we dubbed VIPRnet, shown in Figure 5.

```
VIPRnet(
  (conv_layers): Sequential(
    (0): Conv2d(1, 32, kernel_size=(3, 3), stride=(1, 1), padding=(1, 1))
    (1): ReLU()
    (2): MaxPool2d(kernel_size=2, stride=2, padding=0, dilation=1, ceil_mode=False)
    (3): Conv2d(32, 64, kernel_size=(3, 3), stride=(1, 1), padding=(1, 1))
    (4): ReLU()
    (5): MaxPool2d(kernel_size=2, stride=2, padding=0, dilation=1, ceil_mode=False)
    (6): Conv2d(64, 128, kernel_size=(3, 3), stride=(1, 1), padding=(1, 1))
    (7): ReLU()
    (8): MaxPool2d(kernel_size=2, stride=2, padding=0, dilation=1, ceil_mode=False)
  )
  (fc_layers): Sequential(
    (0): Flatten(start_dim=1, end_dim=-1)
    (1): Linear(in_features=131072, out_features=128, bias=True)
    (2): ReLU()
    (3): Dropout(p=0.5, inplace=False)
    (4): Linear(in_features=128, out_features=1, bias=True)
  )
)
```

Figure 5: VIPRnet model structure

# 3. Results

## 3.1 YOLO Segmentation Model Results

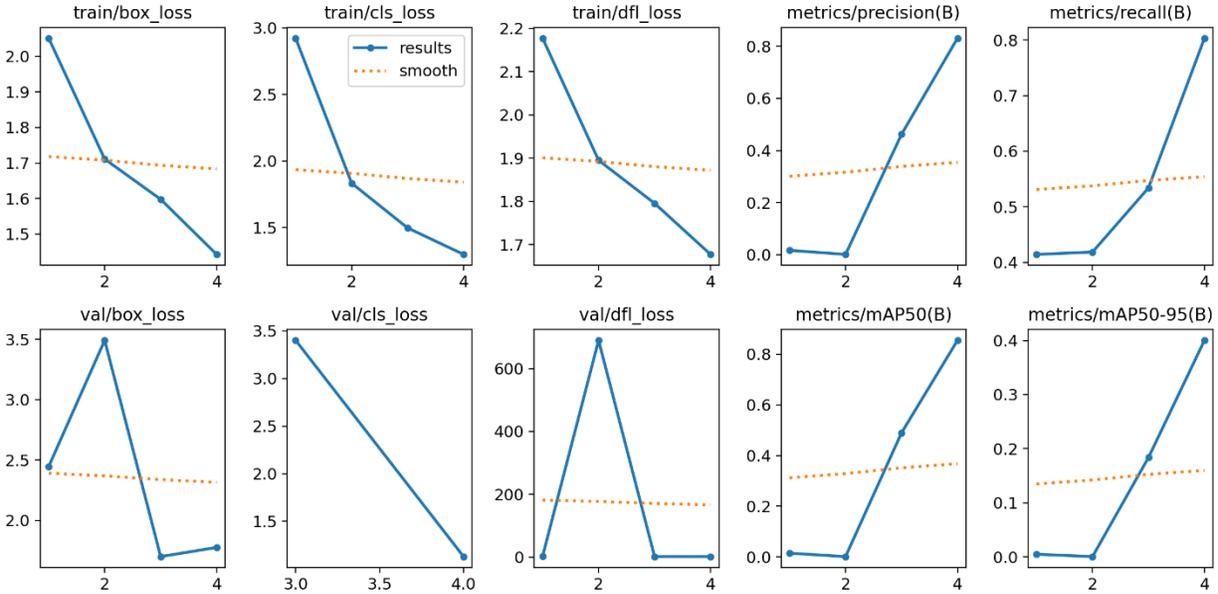

Figure 6: Training and Validation Loss & Metric Curves

### 3.1.1 Training Performance

As shown in Figure 6, the model achieved strong convergence across all loss components. Training loss decreased steadily over 4 epochs:

- **Box loss**: 2.07 → 1.43
- **Classification loss**: 2.94 → 1.43
- **DFL loss**: 2.18 → 1.68

Validation metrics showed a sharp improvement after the initial epoch, with mAP@0.5 rising to 0.78 and mAP@0.5:0.95 reaching 0.40. These improvements are corroborated by increasing precision (0.84) and recall (0.80) values.

| Metric | Value |
|---|---|
| Precision | 0.84 |
| Recall | 0.80 |
| mAP@0.5 | 0.78 |
| mAP@0.5:0.95 | 0.40 |

Table 2: YOLO Segmentation model training metrics

### 3.1.2 Confidence-Based Analysis

The Recall-Confidence curve (Figure 7) demonstrates strong recall even at low thresholds, suggesting the model rarely misses true positives. Conversely, the Precision-Confidence curve (Figure 8) indicates that perfect precision (1.00) is only achieved at a high threshold (0.952), resulting in reduced recall. The F1-confidence curve (Figure 9) peaks at 0.701, offering an optimal trade-off between the two metrics for general-purpose detection.

In the context of this research, the F1-optimal threshold of 0.701 was selected rather than a precision-maximizing threshold to prioritize sensitivity in detecting potential vocal cord anomalies. In a clinical screening workflow, this ensures that fewer pathological cases are missed.

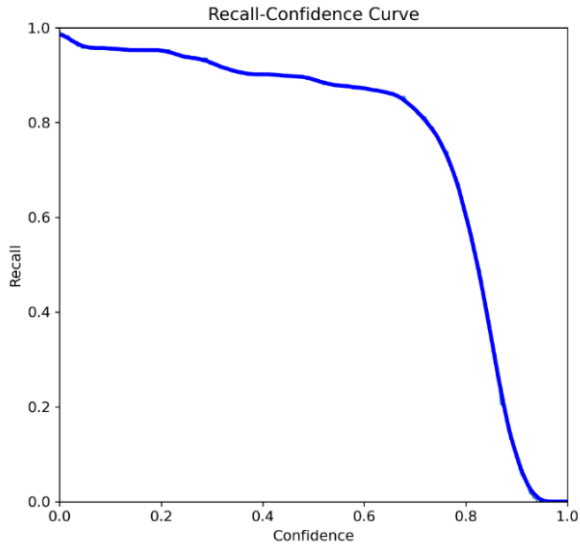
Figure 7: Recall-Confidence curve

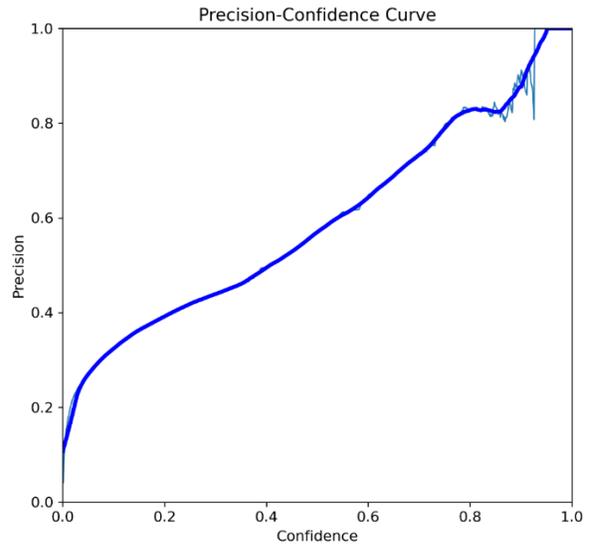
Figure 8: Precision-Confidence Curve

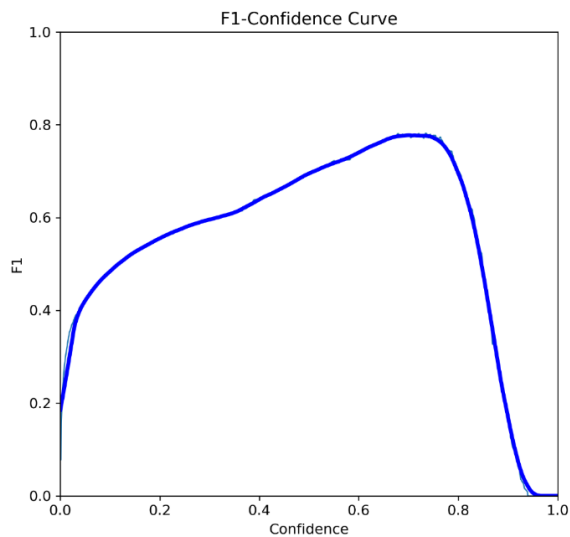
Figure 9: F1-Confidence Curve

### 3.1.3 Spatial and Dimensional Consistency

As illustrated in Figures 10 and 11, bounding boxes predominantly cluster near the image center (x ≈ 0.5, y ≈ 0.4), aligning with anatomical expectations. A strong correlation between box width and height was observed, corresponding to the elongated shape of vocal cords.

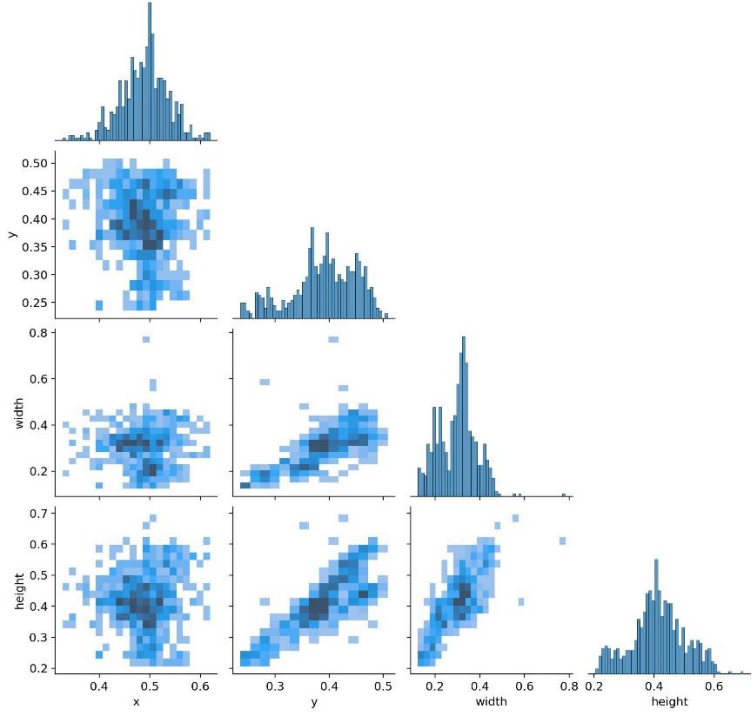

Figure 10: Label Distribution Grid (Instances, x, y, width, height)

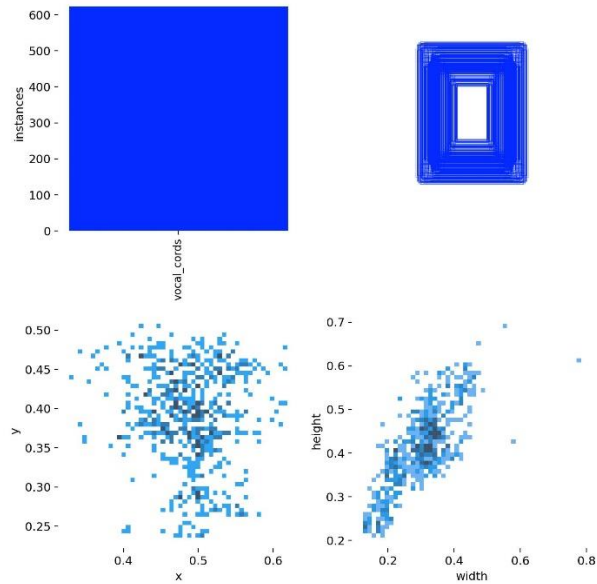

Figure 11: Pairwise Correlogram of Bounding Box Parameters

### 3.1.4 Confusion Matrix

The normalized confusion matrix (Figure 12) reveals strong performance on the single-class task, with 96% accuracy in identifying vocal cords with minimal misclassification of vocal cords as background.

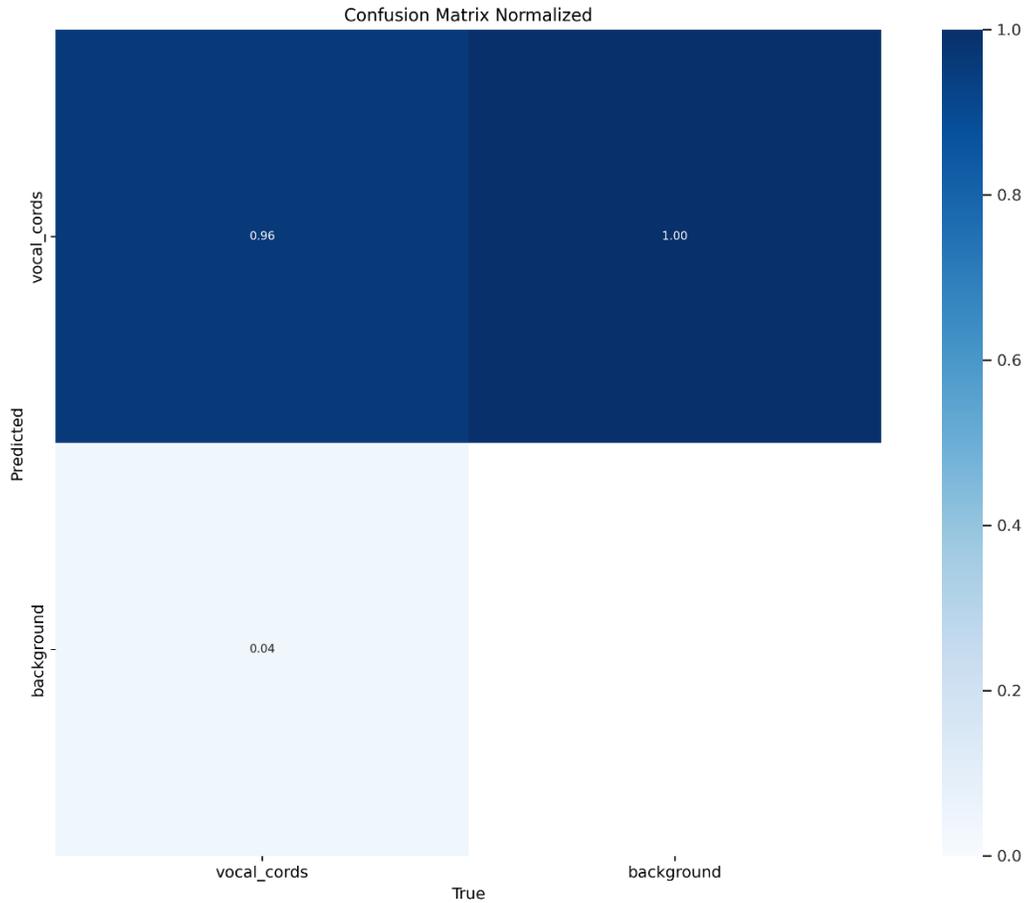

Figure 12: Normalized Confusion Matrix

### 3.1.5 Qualitative Results

Figure 13 presents bounding box predictions on representative validation images. The model consistently identified vocal cords even under conditions of shadowing or simulated asymmetry, reinforcing its robustness across imaging artifacts and augmentations.

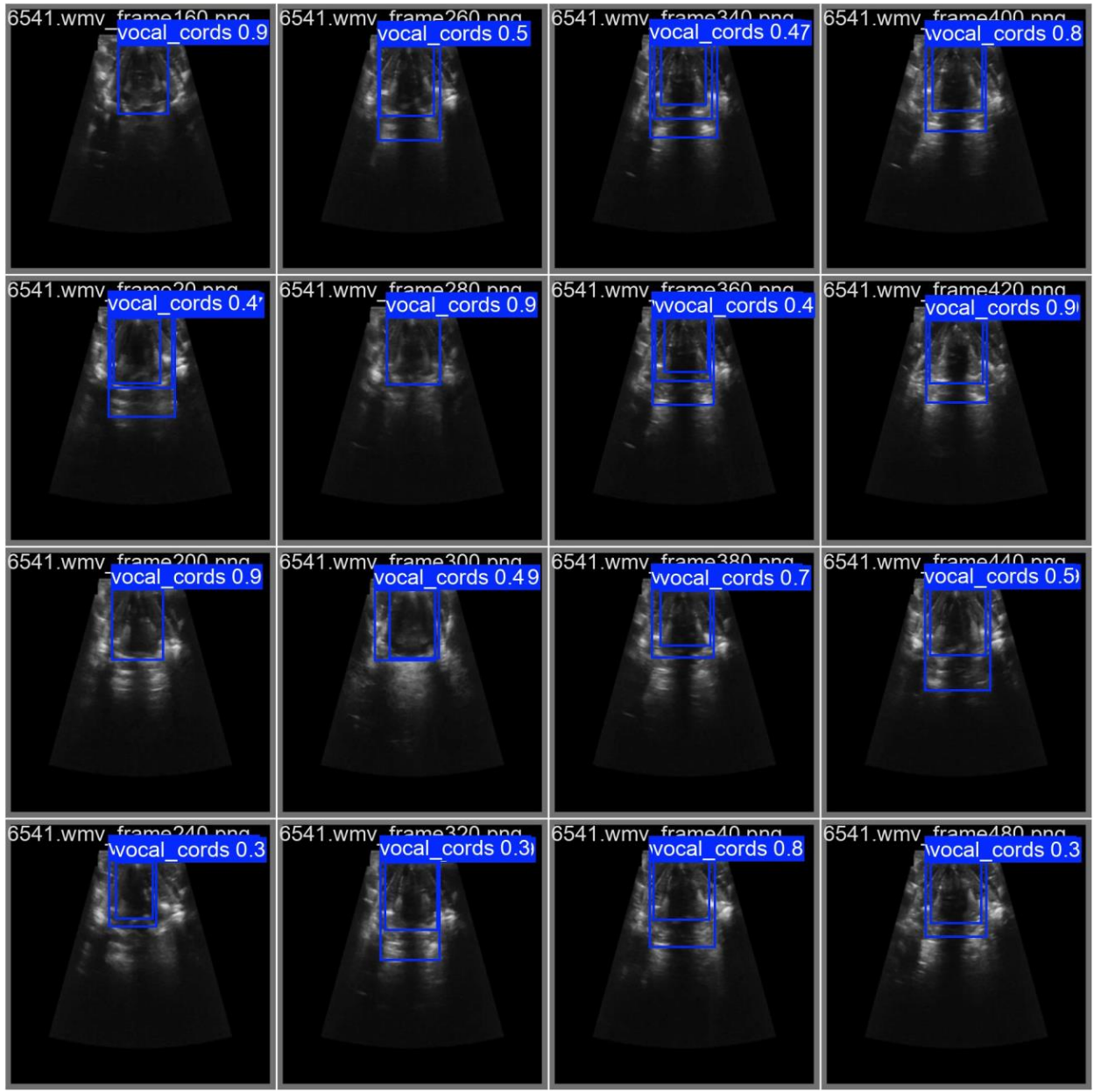

Figure 13: Sample Detection with Bounding Boxes on Ultrasound Image

## 3.2 YOLO Classification Model Results

### 3.2.1 Training Performance

The YOLOv8n-cls model converged reliably over 20 training epochs, with both training and validation loss showing a consistent downward trend (Figures 14a–b). Top-1 accuracy increased from 0.48 to 0.93, indicating strong learning of class-discriminative features. Top-5 accuracy reached 1.00 early in training and remained flat, reflecting the binary nature of the classification task (Figures 14c–d).

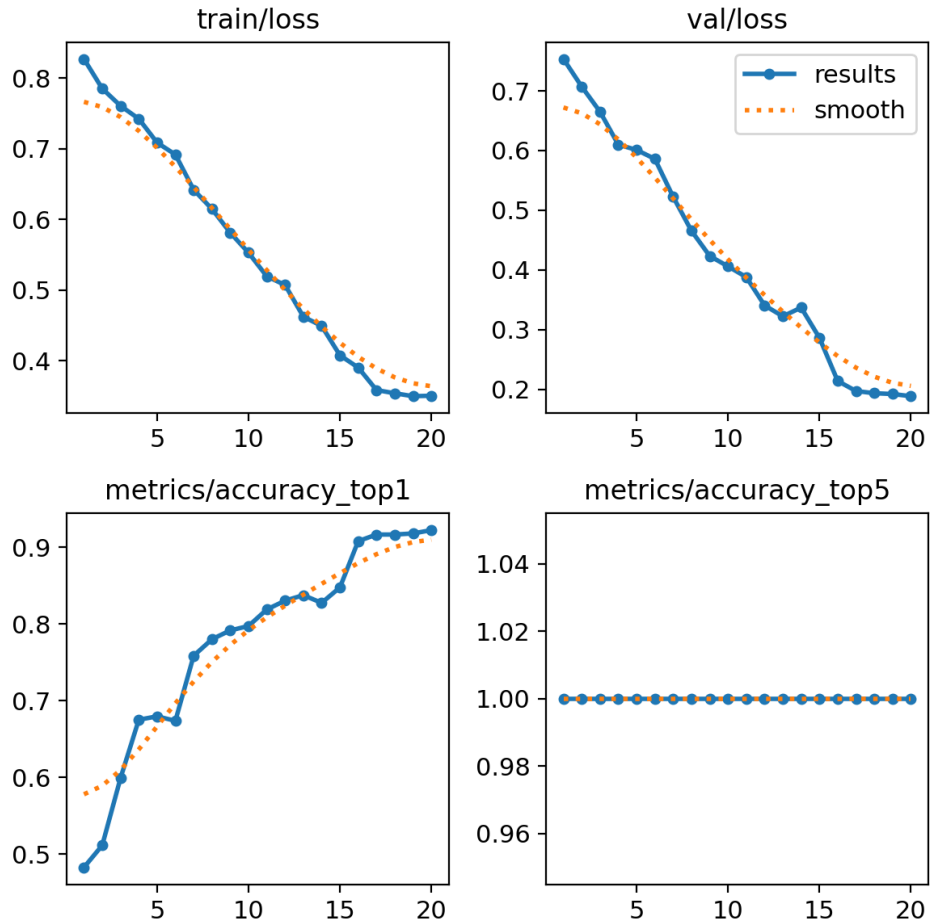

Figure 14: Training and Validation Curves for Classification Model
(a) Training Loss    (b) Validation Loss    (c) Top-1 Accuracy    (d) Top-5 Accuracy

### 3.2.2 Classification Accuracy

On the held-out validation set, the classification model achieved a final top-1 accuracy of 92.3%. As illustrated in the normalized confusion matrix (Figure 15), the model correctly identified 87% of *healthy* cases and 97% of *paralyzed* cases. Most misclassifications were false positives—classifying healthy cords as paralyzed—an acceptable tradeoff in clinical screening contexts where recall is prioritized.

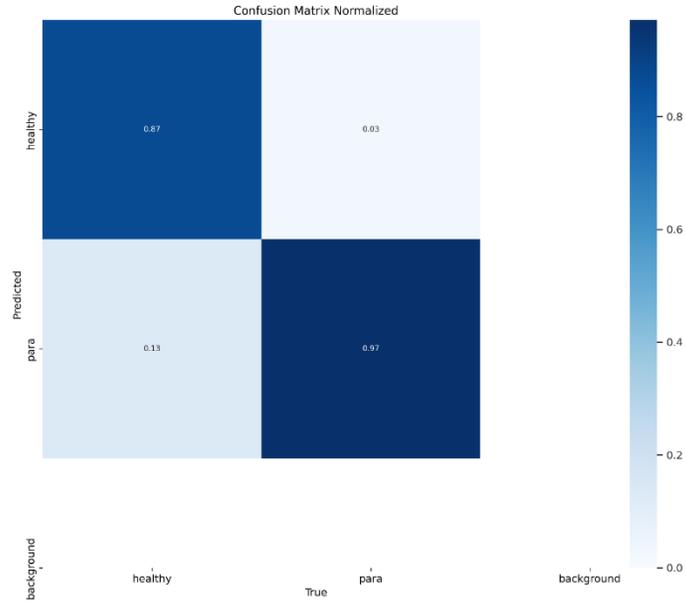

Figure 15: Normalized Confusion Matrix for Vocal Cord Classification

To further evaluate classifier performance under different thresholds, a precision-recall (PR) curve was plotted (Figure 16). The curve shows high precision across a wide recall range, with a steep drop-off only at extreme recall values. This shape confirms the model's reliability in detecting true positives with minimal false positives under moderate thresholds—an essential characteristic for early-stage screening tools in clinical settings.

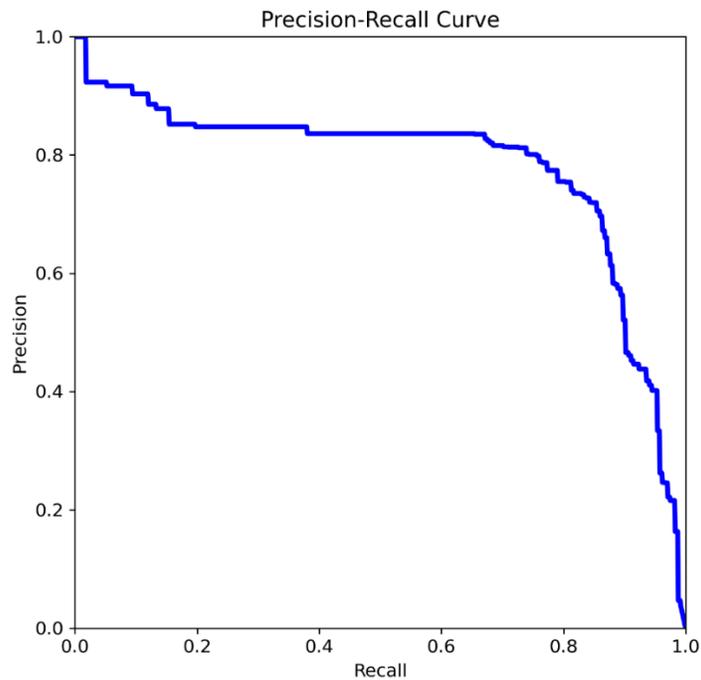

Figure 16: Precision-recall curve for the classification task. The area under the curve and flatness of the plateau suggests the model maintains high precision even with increasing recall.

## 3.3 VIPRnet Classification Model Preliminary Results

Training performance for VIPRnet is presented in Figure 17. VIPRnet was trained over 50 epochs with a batch size of 64, showing a gradual and consistent reduction in both training and validation loss. VIPRnet exhibited a smooth learning curve with minimal indicators of overfitting, as evidenced by the small gap between training and validation metrics. By the final epoch, the model stabilized at approximately 99.5% validation accuracy, indicating strong performance with our synthetic data. Further analysis was unfortunately not possible due to time constraints.

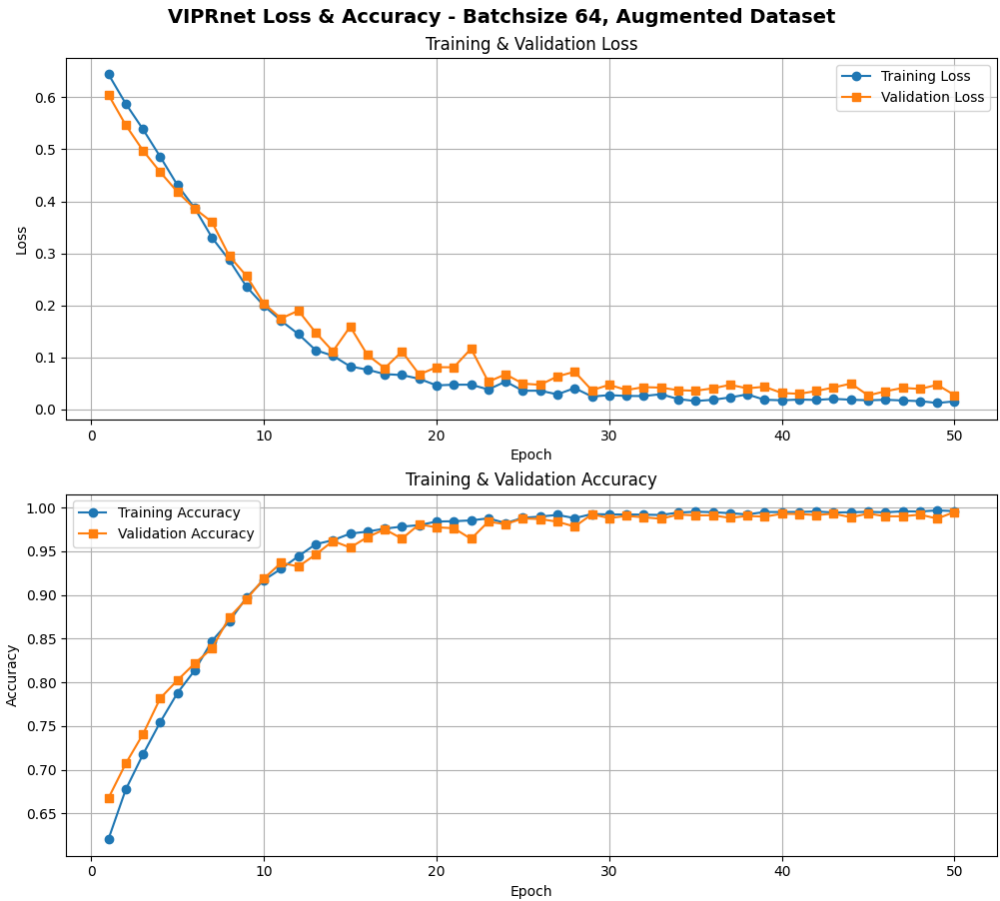

Figure 17: VIPRnet training metrics

## 3.4 Results Summary

Despite differences in their training durations and batch sizes, both architectures achieved effective convergence. Furthermore, both models exhibit extremely high validation accuracies, which gives promise that they will perform adequately at the task they were trained for (differentiating VCUS images acquired from a small population of healthy undergraduate students apart from digitally modified versions of those same images). However, these results necessarily reflect the limitations of our study and should *not* be interpreted as a guarantee that either model will generalize to a clinical setting or exhibit these same levels of accuracy on VCUS images featuring genuine VCP anatomy.

## 4. Discussion

VCUS may be a preferable method to look for VCP as compared to flexible laryngoscopy (FL) because it is non-invasive, cost-effective, quick, well-tolerated with no patient discomfort, and accurate as reported by experts in the technique. However, it is widely perceived to be difficult to perform or interpret by those less familiar with the technique, with reported operator variability as a limitation of VCUS [8,9]. To mitigate human operator-dependence, machine learning models have the potential to provide consistently accurate analysis [13].

This study demonstrates the feasibility of using machine learning to detect vocal cord paralysis (VCP) via VCUS. By recording ultrasound videos from 30 participants, parsing the footage into still frames, and manually labeling ROIs containing the vocal cords, we were able to train a YOLOv8 based segmentation model that segmented the cords with high accuracy. Further, we explored the viability of multiple convolutional neural networks such as the pretrained YOLOv8n-cls and our custom VIPRnet architecture to classify images as healthy or paralyzed.

To address the lack of a clinical source of VCUS images showcasing true VCP, we generated synthetic VCP frames by "asymmetrically squishing" one half of each labeled ROI and blending the halves back with bilinear interpolation. This technique, combined with data augmentation, substantially expanded our training set to 34,816 unique images, while also mitigating class imbalance. Although our synthetic approach cannot fully replicate complex pathological variations and potentially caused model overfitting, it enabled us to rigorously train and tune several models with promising preliminary results.

Looking ahead, we plan to refine our approach and expand our datasets in several ways. The most obvious next step is to acquire clinical VCP data to validate the models we have developed and potentially train new ones. Barring that, a more robust use of YOLO segmentation could allow us to automatically identify and label a larger number of frames from our current and future IRB-approved data collections, thereby increasing the volume and diversity of available training images. Additional hyperparameter tuning, improved synthetic data generation techniques, and exploration of alternate classification architectures could also further boost model performance. Additionally, we would like to investigate the methodology of a recently published videolaryngoscopy-based study [13], borrowing their length measurement framework and applying it to our ultrasound dataset. Finally, as our work continues to evolve, we seek to collaborate both with other researchers to create more clinically useful tools and models, and with private medical device manufacturers to potentially incorporate our algorithms into commercial ultrasound hardware to run in real-time at the point of care.

## 5. Conclusion

Machine learning-assisted analysis of VCUS shows great promise in improving diagnostic accuracy over operator-dependent human interpretation.

## Acknowledgements

We would like to extend our thanks Dr. Jeffrey LaMack for his unwavering support, brainstorming, advice, and assistance as our team's faculty supervisor. We would also like to thank Dr. Derek Riley for his suggestions and ideas related to how we technically approached the problems of VCP classification and synthetic data generation. Lastly, we would like to thank Dr. Kristen Shebesta for her patience and assistance in securing IRB approval for our work.